\documentclass[10pt,twocolumn,letterpaper]{article}

\usepackage{cvpr}
\usepackage{times}
\usepackage{epsfig}
\usepackage{graphicx}
\usepackage{amsmath}
\usepackage{amssymb}

\usepackage{url}            
\usepackage{booktabs}       
\usepackage{MnSymbol}
\usepackage{amsfonts}       
\usepackage{gensymb}		
\usepackage{nicefrac}       
\usepackage{microtype}      
\usepackage{color}			
\usepackage[labelfont={bf,small},font=small]{caption}
\usepackage[labelfont={bf,small},font=small]{subcaption}
\usepackage{multirow}
\usepackage{makecell}
\usepackage{lipsum}
\usepackage{comment}

\usepackage{soul}           

\usepackage{framed}

\makeatletter
\renewcommand{\paragraph}{%
  \@startsection{paragraph}{4}%
  {\z@}{0.75ex \@plus 1ex \@minus .2ex}{-1em}%
  {\normalfont\normalsize\bfseries}%
}
\makeatother

\def\mbf#1{\mathbf{#1}}

\setcellgapes{3pt}

\newcolumntype{?}[1]{!{\vrule width #1}}

\newlength{\Oldarrayrulewidth}

\usepackage[breaklinks=true,bookmarks=false,colorlinks=true]{hyperref}

\cvprfinalcopy 
\def\cvprPaperID{6318} 

\begin{document}

\title{AIRD: Adversarial Learning Framework for Image Repurposing Detection}

\author{Ayush Jaiswal, Yue Wu, Wael AbdAlmageed, Iacopo Masi, Premkumar Natarajan\\
USC Information Sciences Institute, Marina del Rey, CA, USA\\
{\tt\small \{ajaiswal, yue\_wu, wamageed, iacopo, pnataraj\}@isi.edu}
}

\maketitle


\begin{abstract}
    Image repurposing is a commonly used method for spreading misinformation on social media and online forums, which involves publishing untampered images with modified metadata to create rumors and further propaganda. While manual verification is possible, given vast amounts of verified knowledge available on the internet, the increasing prevalence and ease of this form of semantic manipulation call for the development of robust automatic ways of assessing the semantic integrity of multimedia data. In this paper, we present a novel method for image repurposing detection that is based on the real-world adversarial interplay between a bad actor who repurposes images with counterfeit metadata and a watchdog who verifies the semantic consistency between images and their accompanying metadata, where both players have access to a reference dataset of verified content, which they can use to achieve their goals. The proposed method exhibits state-of-the-art performance on location-identity, subject-identity and painting-artist verification, showing its efficacy across a diverse set of scenarios.
\end{abstract}



\section{Introduction}
\label{sec:introduction}

The internet-driven information age, in which we are currently living, has seen rapid advances in technology that have made creation and transmission of information on a large-scale increasingly easier. Consequently, the manner in which people consume information has evolved from printed media and cable-television to digital sources. Simultaneously, social networking platforms have evolved to make it easier for people to disseminate information quickly within communities and publicly. This provides an excellent way for people to share news quickly, making social media a popular news source, especially among the youth~\cite{bib:marchi2012}. However, this ease of information propagation has also made social networks a popular mode of transmission of fake news.

\begin{figure}[t]
\captionsetup[figure]{aboveskip=3pt,belowskip=-5pt}
\captionsetup[subfigure]{aboveskip=-2pt,belowskip=-5pt}
\centering
\begin{subfigure}{\linewidth}
\centering
\includegraphics[width=0.42\linewidth]{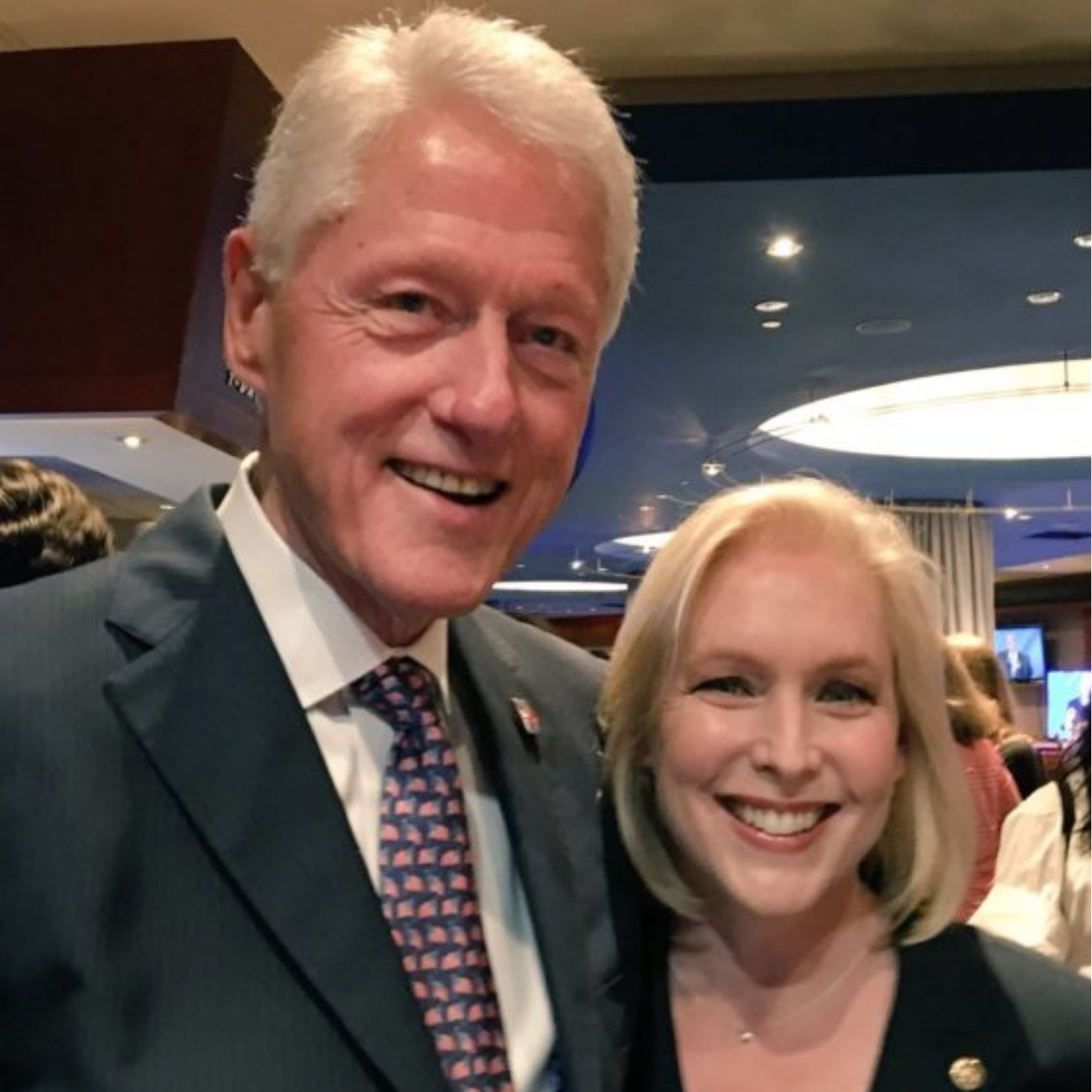}
\hspace{10pt}
\includegraphics[width=0.42\linewidth]{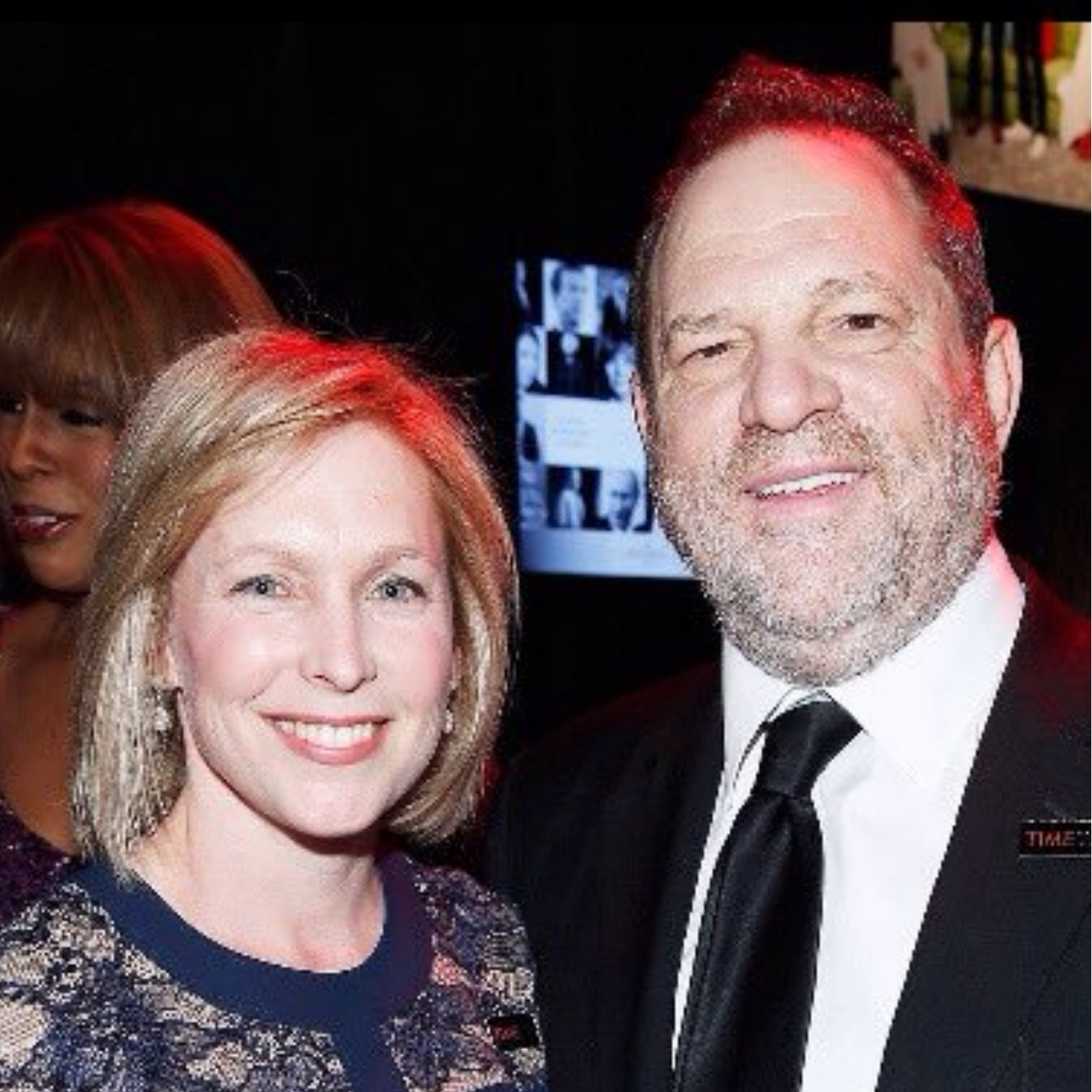}
\includegraphics[height=0.5cm,trim={0 0 0 0.25cm},clip]{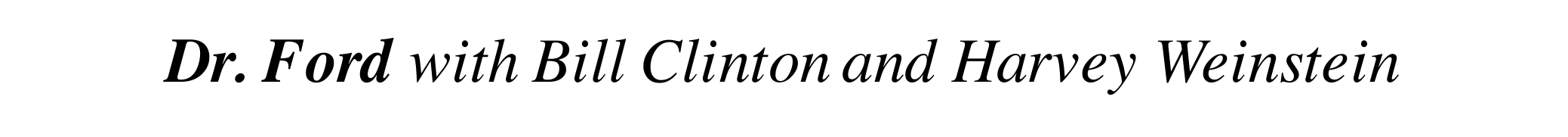}
\caption{Subject-Identity\label{fig:new_real_repurposing_examples:faces}}
\end{subfigure}
\par\medskip
\begin{subfigure}{0.42\linewidth}
\centering
\includegraphics[width=\linewidth]{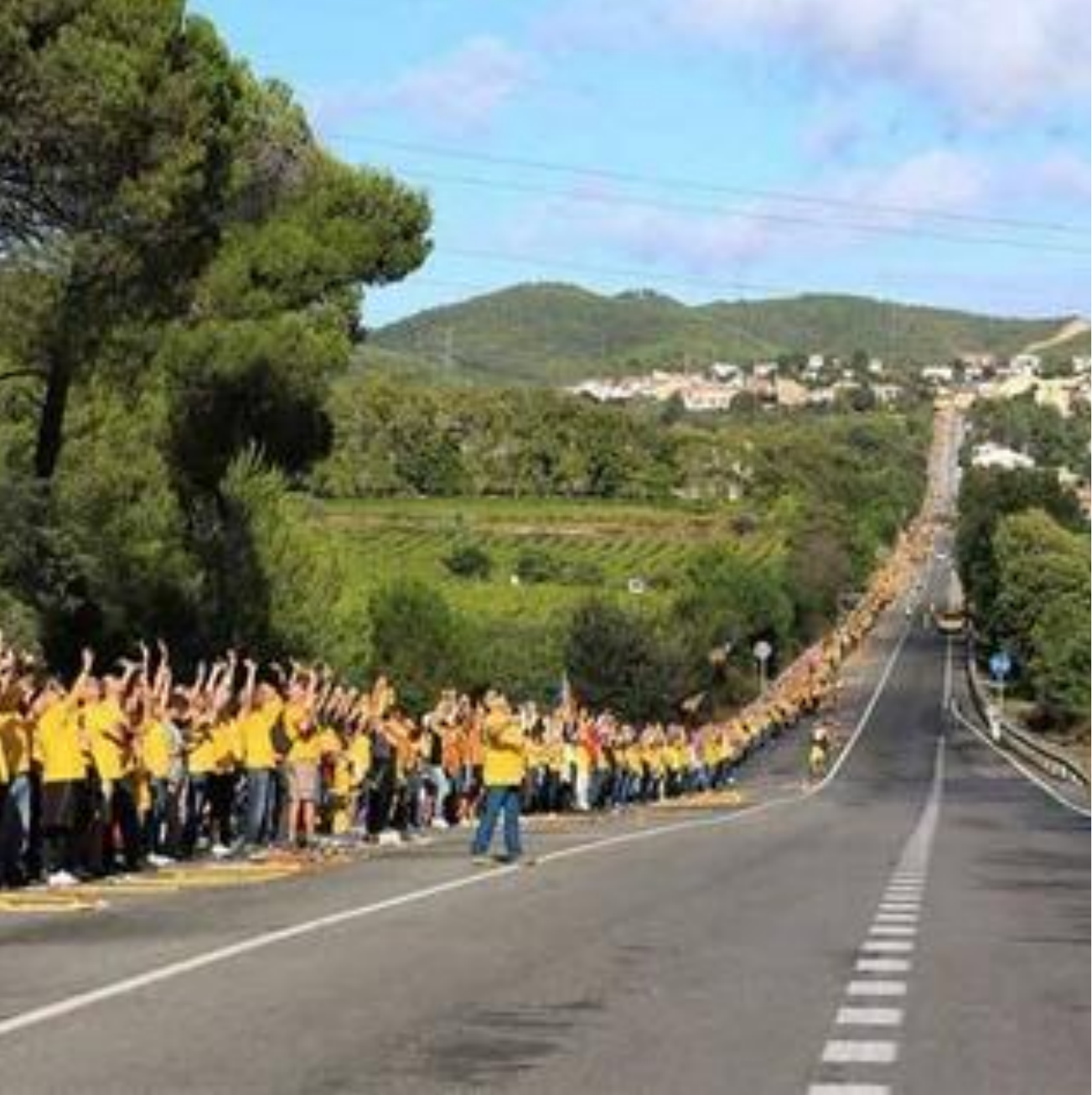}
\includegraphics[height=0.5cm,trim={1.25cm 0 1.25cm 0.25cm},clip]{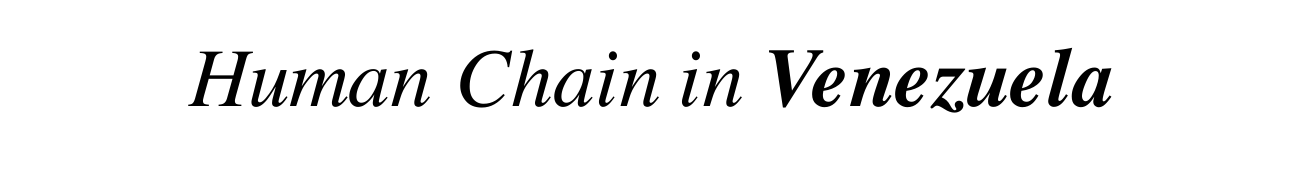}
\caption{Location-Identity\label{fig:new_real_repurposing_examples:locations}}
\end{subfigure}
\hspace{10pt}
\begin{subfigure}{0.42\linewidth}
\centering
\includegraphics[width=\linewidth]{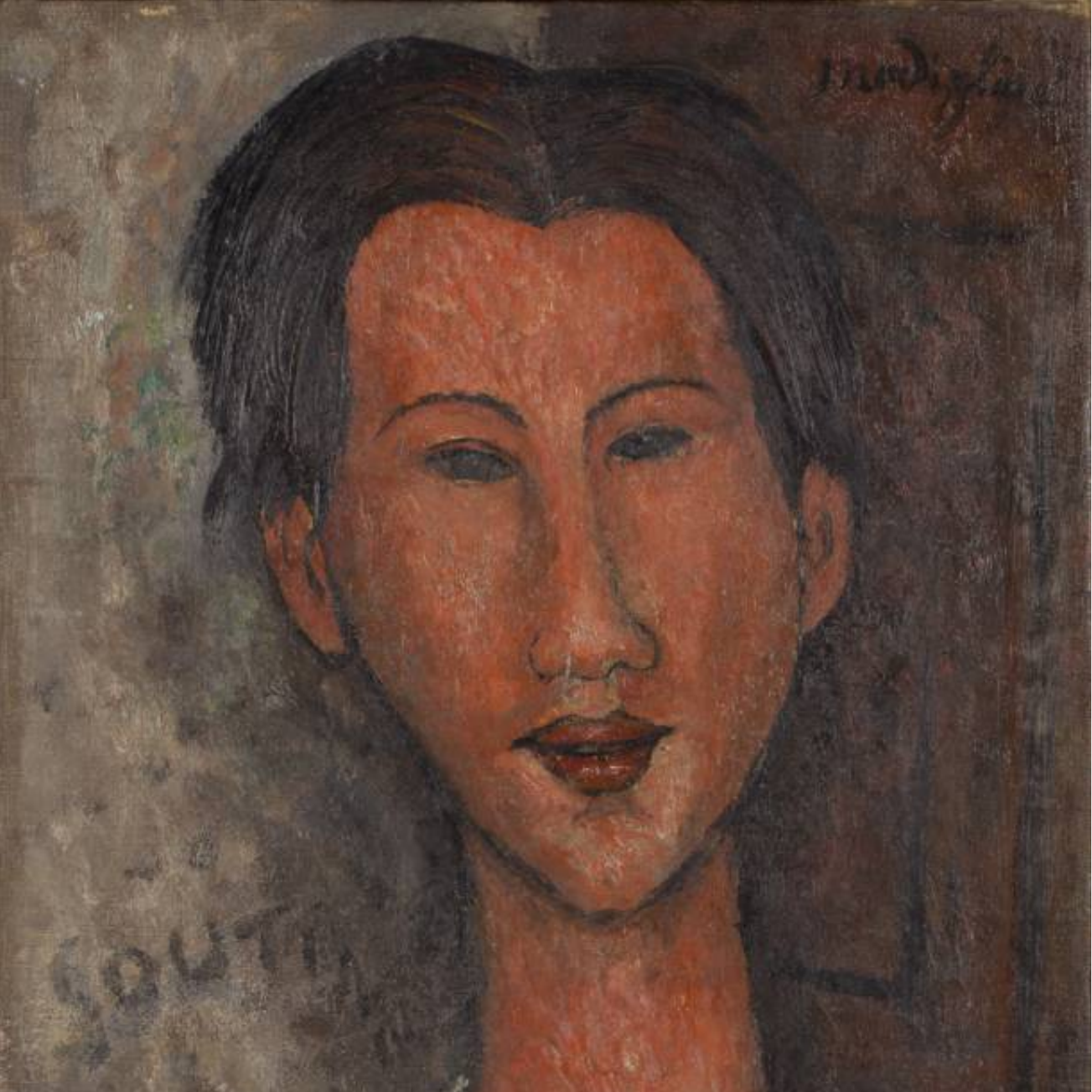}
\includegraphics[height=0.5cm,trim={1.25cm 0 1.25cm 0.25cm},clip]{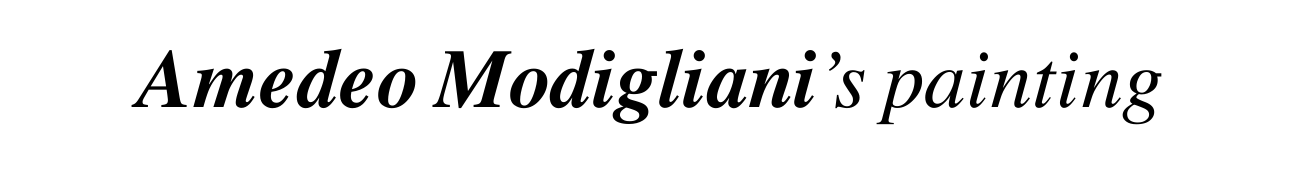}
\caption{Painting-Artist\label{fig:new_real_repurposing_examples:paintings}}
\end{subfigure}
\caption{\label{fig:teaser}Image repurposing in different domains from real examples -- (a) lookalike\protect\footnotemark, (b) incorrect location\protect\footnotemark, (c) wrongly-claimed artist\protect\footnotemark. Unmanipulated images are often reused in this way to spread misinformation about a similar yet different entity or event.}
\end{figure}

\addtocounter{footnote}{-3}

Given the potency of falsified information propagating on the internet, several activist groups have launched crowd-sourced efforts\stepcounter{footnote}\footnotetext{\label{fn:face}\href{https://twitter.com/jfnyc1/status/1043665540580081669}{https://twitter.com/jfnyc1/status/1043665540580081669}}\stepcounter{footnote}\footnotetext{\label{fn:loc}\href{https://twitter.com/tachirense89/status/434844811548311552}{https://twitter.com/tachirense89/status/434844811548311552}}\stepcounter{footnote}\footnotetext{\label{fn:painting}\href{https://bit.ly/2QLJjaC}{https://bit.ly/2QLJjaC}} (such as Snopes\footnote{\href{https://www.snopes.com/}{www.snopes.com}}) towards debunking fake news. However, the sheer volume and rate at which information is being created and disseminated necessitate developing automated ways of validating information. Several methods have, hence, been developed recently for detecting rumors on online forums~\cite{bib:gupta2012,bib:jin2013,bib:liu2015,bib:ma2016,bib:wu2015wu,bib:ruchansky2017}, digital manipulations of images~\cite{bib:asghar_copy-move_2017,wu2017deep_acmm,wu2018busternet,wu2018image}, and semantic inconsistencies in multimedia data~\cite{bib:jaiswal2017,bib:sabir2018}. While the detection of digital manipulation has gained most of the research attention over the years, rumor detection and semantic integrity verification are much newer areas of research.

In this paper, we focus on detecting image repurposing --- a form of semantic manipulation of multimedia data where \emph{untampered} images are \emph{reused} in combination with falsified metadata to spread misinformation. Figure~\ref{fig:teaser} shows some real-world examples of image repurposing. Jaiswal~\emph{et al.}~\cite{bib:jaiswal2017} define the broader problem of multimedia semantic integrity assessment as the verification of consistency between the media asset (e.g. image) and different components of the associated metadata (e.g. text caption, geo-location, etc.), since the asset and the metadata are expected to be a coherent \textit{package}. They also introduce the concept of using a reference dataset (RD) of untampered packages to assist the validation of \textit{query packages}. Image repurposing detection falls under this umbrella and has been explored in packages of images and captions~\cite{bib:jaiswal2017} as well as those that additionally contain Global Positioning System (GPS) information~\cite{bib:sabir2018}.

The method proposed in~\cite{bib:jaiswal2017} for integrity assessment detects inconsistencies in packages with entire captions potentially copied from other packages at random. Sabir~\emph{et al.}~\cite{bib:sabir2018}, on the other hand, present a method for the detection of manipulations of named entities within captions.  However, the MEIR dataset proposed and evaluated on in~\cite{bib:sabir2018} falls short on the deceptive potential of entity-manipulations because they are implemented as randomly swapping the entity in a given caption with the same class of entity (person, organization, or location) from a caption in an unrelated package.

One of the main challenges for developing image repurposing detection methods is the lack of training and evaluation data. While crowd sourcing is a potential alternative, it is expensive, and time consuming. In light of this, we propose a novel framework for image repurposing detection, \emph{which can be trained in the absence of training data containing manipulated metadata}. The proposed framework, which we call Adversarial Image Repurposing Detection (AIRD), is modeled to simulate the real-world adversarial interplay between a bad  actor who repurposes images with counterfeit metadata and a watchdog who verifies the semantic consistency between images and their accompanying metadata. More specifically, AIRD consists of two models: a counterfeiter and a detector, which are trained adversarially.

Following the approach of previous works, the proposed framework employs a reference dataset of unmanipulated packages as a source of world knowledge. While the detector gathers evidence from the reference set, the counterfeiter exploits it to conjure convincingly deceptive fake metadata for a given query package. The proposed framework can be applied to all forms of metadata. However, since generating natural language text is an open research problem, the experimental evaluation is performed on structured metadata. Furthermore, previous methods on image repurposing detection focus only on entity manipulations within captions. Hence, AIRD could be employed in such cases by first extracting entities using named entity recognition. The proposed framework exhibits state-of-the-art performance on the Google Landmarks dataset~\cite{noh2017largescale} for location-identity verification, a variant of the IJB-C dataset~\cite{maze2018iarpa}, called IJBC-IRD, which we created for subject-identity verification, and the Painter by Numbers dataset~\cite{painter_by_numbers} for painting-artist verification. Results on this diverse collection of datasets, which we make publicly available\footnote{\href{https://github.com/isi-vista/AIRD-Datasets}{www.github.com/isi-vista/AIRD-Datasets}}, illustrate the generalization capability of the proposed model.

The main contributions of this paper are:
\begin{itemize}
    \item a novel approach to image repurposing detection that is modeled to simulate the real-world adversarial interplay between nefarious actors and watchdogs
    \item a new framework design that can be adopted in developing real-world image repurposing detection systems that utilize knowledge-bases to validate information
    \item the IJBC-IRD dataset of face images with subject-identity metadata, created to further research in the area of face-image repurposing detection
\end{itemize}

The rest of the paper is organized as follows. Section~\ref{sec:related_work} discusses related work. In Section~\ref{sec:method} we describe the proposed framework. Results of experimental evaluation are provided in Section~\ref{sec:evaluation}. Finally, we conclude the paper and discuss future directions in Section~\ref{sec:conclusion}.


\section{Related Work}
\label{sec:related_work}
Detection of fake news and rumors on online platforms has been studied in the research domain of rumor detection through automated analysis of textual content~\cite{bib:gupta2012,bib:liu2015,bib:ma2016}, propagation of posts within communities~\cite{bib:jin2013,bib:wu2015wu}, and the kind of response they elicit from people~\cite{bib:ruchansky2017}. These works are not targeted at image-based manipulations of information and do not incorporate any form of image analysis in their methodology of information verification.

Detection of digital manipulations in images has been studied extensively in the past for image-splicing, copy-move forgery, resampling and retouching of images at the pixel-level~\cite{bib:asghar_copy-move_2017,wu2017deep_acmm,wu2018busternet,wu2018image}. These methods work by either verifying embedded watermarks within images or analyzing pixel-level consistencies in search for artifacts.

The reuse of unmanipulated images to spread misinformation about a possibly unrelated entity or event was introduced in~\cite{bib:jaiswal2017} as the verification of semantic integrity of data with image assets. Image repurposing with manipulated textual and location data has been studied more specifically in~\cite{bib:sabir2018}. Our work falls in this category and we propose a novel generalized framework for image repurposing detection.

Adversarial learning has been employed recently for improving object detection~\cite{bib:zhang2018}, disentangled feature learning~\cite{bib:jaiswal2018,bib:liu2018} and feature augmentation~\cite{bib:volpi2018}, besides data generation~\cite{bib:capsgan,bib:bicogan}. The proposed AIRD framework uses adversarial learning to model the real-life interplay between malicious actors and watchdogs for image repurposing detection.


\section{Adversarial Image Repurposing Detection}
\label{sec:method}

\begin{figure}
\centering
\captionsetup{aboveskip=3pt}
\includegraphics[width=\linewidth,trim={2.3cm 0.02cm 2.25cm 0.02cm},clip]{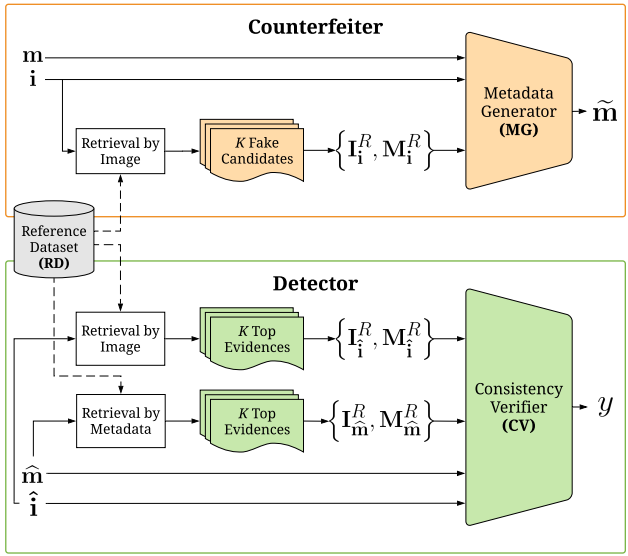}
\caption{\label{fig:aird}Adversarial Image Repurposing Detection (AIRD) --- At the core of this framework are two adversarially trained models: the Counterfeiter $\mathcal{C}$ and the Detector $\mathcal{D}$. Reference Dataset (RD) is a collection of verified images and metadata representing world knowledge. Variants of $\mbf{i}$ and $\mbf{m}$ are used to denote images and metadata, respectively. Similarly, variants of $\mbf{I}^R$ and $\mbf{M}^R$ denote collections of retrieved images and metadata, respectively. While the Metadata Generator (MG) of $\mathcal{C}$ takes advantage of RD to conjure fake metadata for an image by analyzing that of other similar images \emph{of different entities or events}, the Consistency Verifier (CV) of $\mathcal{D}$ uses evidence gathered from the RD to assess the veracity of the claimed metadata presented with the query image.}
\end{figure}

Jaiswal~\emph{et al.}~\cite{bib:jaiswal2017} introduced the general approach of using a reference dataset of packages as a knowledge-base from which evidence could be gathered, to assist the semantic integrity assessment of a query package. Conceptually, this approach is similar to how watchdogs verify news articles online, with news sources, document repositories and encyclopedias on the internet serving as enormous reference datasets. However, in the real-world, these datasets are accessible to nefarious actors too. In the case of image-repurposing, reference datasets are exploited in order to find images that can serve as fake evidences for rumors and propaganda. Thus, both counterfeiters and watchdogs have access to the same information but one group uses it for spreading misinformation while other employs it for information validation. The presented view of this key observation reveals an inherent \emph{informed} adversarial interplay between the two groups. We propose the Adversarial Image Repurposing Detection (AIRD) framework that models this interplay.

The proposed AIRD framework consists of two competing models --- (1) a counterfeiter ($\mathcal{C}$), which uses the reference dataset to fabricate metadata for untampered images, and (2) a detector ($\mathcal{D}$), which gathers evidence from the reference dataset  to verify the semantic integrity of query packages. While the working mechanism of the detector network is close to that of real-world watchdogs, we model the counterfeiter as an extremely malevolent person who wishes to repurpose all available images with the sole intention of spreading as much misinformation as possible. At the core of the counterfeiter model is a metadata-generator (MG) neural network, while the detector contains a consistency-verifier (CV) network. The parameters of these networks are learned through adversarial training. The models ($\mathcal{C}$ \& $\mathcal{D}$) employ semantic encoding and retrieval tools to fetch additional packages from the reference set for their individual goals. Figure~\ref{fig:aird} shows the high-level design of the proposed framework. We describe the components of AIRD including implementation and training details in the following sections.

\subsection{Encoding and Retrieval from Reference Dataset}
\label{subsec:encoders}

\paragraph{Modality-specific encoders:} The counterfeiter and detector models in the proposed framework gather information through retrieval of additional similar and/or related packages from the reference dataset. In order to facilitate meaningful retrieval of information, it is important to encode each modality of the multimedia packages into information-rich semantic representations. Ideally, these encoders would be trained end-to-end with the metadata-generator and consistency-verifier networks, so that each can extract very specific information and learn similarity between data instances in order to help it achieve its goal. For example, in the case of packages containing images and captions, both MG and CV would have their own copies of image and text encoders that would be trained end-to-end with them. However, this is not always feasible because the entire reference set, which is enormous in size, would have to be re-indexed every time the parameters of these encoders are updated for complex modalities like images and text.

The proposed framework employs pretrained modality-specific encoders that are not updated during the adversarial training, when such situations arise. It is crucial to carefully select these encoders such that the embeddings generated by them capture all details of information that is vulnerable to manipulation. Thus, an ideal system would use image and text encoders that capture fine-grained semantic details in images and captions, in our example above. As discussed earlier, the proposed framework is evaluated on packages containing images and highly-specific structured metadata. Hence, we employ off-the-shelf yet state-of-the-art deep neural networks trained externally to detect those metadata as image encoders. In contrast, the metadata encoders are learned jointly with the adversarial training of MG and CV, such that the models can learn to cluster the metadata values by similarity  in the embedding space. In the rest of the paper, we denote the encodings of image and metadata (whether natural language or structured) as $\mbf{i}$ and $\mbf{m}$, respectively.

\paragraph{Indexing and retrieval from the reference dataset:} Retrieval of additional related information from the reference dataset based on structured metadata is naturally implemented as database querying. Contrarily, data modalities like images and text have to be first encoded into vector representations. Retrieving related packages can then be performed through a nearest neighbor search. However, in the case of semantic integrity assessment, the reference dataset is expected to be enormous in size, especially in real world-cases, where it is expected to contain possibly all validated knowledge about the world. Using a brute-force nearest neighbor search, therefore, becomes impractical.

In order to develop a scalable framework for image repurposing detection, we employ efficient approximate search methods for similarity-based querying of reference datasets that have been shown to work on databases of billions of records. Furthermore, to boost the accuracy of the approximation, we use a cascaded indexing mechanism. The high-level stages of the system are --- (1) indexing using a Reranked Product Quantization-based Inverted File Index (IVFPQ+R)~\cite{bib:ivfpqr}, followed by (2) additional re-ranking of approximate retrievals using exact cosine similarities. We use efficient implementations of these indexing modules publicly available in the \texttt{faiss}\footnote{\href{https://github.com/facebookresearch/faiss}{www.github.com/facebookresearch/faiss}} package.

\subsection{Counterfeiter Model $\mathcal{C}$}

The working mechanism of $\mathcal{C}$ is shown in the upper-half of Figure~\ref{fig:aird} and is described as follows.
\paragraph{Fake candidates:} The non-parametric component of $\mathcal{C}$ aims to find plausible misleading candidates. In order to repurpose an image (encoded as $\mbf{i}$) by manipulating its metadata ($\mbf{m}$), the counterfeiter first queries the reference dataset for the $K$-most similar images \emph{with dissimilar metadata}. The encodings of these images are collectively denoted as $\mbf{I}_{\mbf{i}}^R$. Similarly, their accompanying metadata is denoted as $\mbf{M}_{\mbf{i}}^R$. Given the characteristics of modality-specific encoders described in Section~\ref{subsec:encoders}, this results in $K$ images that could be confused for the original with respect to the metadata to be manipulated. For instance, if the image of a person's face were to be repurposed as someone else's by manipulating the subject-identity metadata, such a retrieval would result in $K$ face images whose subjects look very similar to the original subject. We call these retrieved packages \textit{fake candidates}.
\paragraph{Metadata Generator:} The fake candidates as well as the original image and metadata are then passed on to the metadata generator neural network (MG). While caption-metadata is already encoded using a pretrained encoder (as described above), in the case of structured metadata, MG first encodes it using a metadata encoder that is trained as a part of MG. MG contains a candidacy-scorer sub-network (CSSN; implemented using two fully-connected layers), which then scores each of the $K$ candidates by comparing it with the original image-metadata pair, as shown in Equation~\ref{eq:cssn}:

\begin{equation}
\label{eq:cssn}
    s_k = \text{CSSN} \Big ( (\mbf{i}, \mbf{m}), (\mbf{i}_k, \mbf{m}_k) \Big ),
\end{equation}
where $\mbf{i}_k$ and $\mbf{m}_k$ denote the $k$-th package in $\{\mbf{I}_{\mbf{i}}^R$, $\mbf{M}_{\mbf{i}}^R\}$.

Finally, scores of the candidates are converted into a choice distribution ($\mbf{c}$) through an attention-like softmax operation. In order to make the choices sharp, softmax is used with a low temperature, where softmax with temperature, as used in reinforcement learning for converting values into action-decisions, is described in Equation~\ref{eq:softmax_temp}.

\begin{equation}
\label{eq:softmax_temp}
    c_k = \frac{\exp \left ( s_k / \tau \right )}{\sum_{j=1}^{K} \exp \left ( s_j / \tau \right )} \ ; \ \tau \in (0, 1]
\end{equation}

The choice distribution is multiplied element-wise with the metadata of the fake candidates. MG then produces the fabricated metadata as the sum of these weighted candidate metadata. Since the choice distribution is sharp, this simulates the act of choosing one of the $K$ metadata values while still retaining differentiability. The fabricated metadata is, thus, computed as described in Equation~\ref{eq:mfake}.

\begin{equation}
\label{eq:mfake}
    \widetilde{\mbf{m}} = \sum_{k=1}^{K} c_k \cdot \mbf{m}_k
\end{equation}

\subsection{Detector Model $\mathcal{D}$}

The lower-half of Figure~\ref{fig:aird} provides an overview of the working mechanism of $\mathcal{D}$. It is described as follows.

\paragraph{Gathering evidence:} The manual process of gathering evidence from reference datasets and using them to validate query packages inspires the design of the detector model ($\mathcal{D}$) in the proposed AIRD framework. The detector starts with retrieving $K$-most similar packages from the reference dataset, using both the image ($\hat{\mbf{i}}$) and the associated metadata ($\widehat{\mbf{m}}$) as the query modality independently. Thus, it gathers two sets of evidence, which can be broken down into image encodings and metadata as $\left \{ \mbf{I}_{\hat{\mbf{i}}}^R, \mbf{M}_{\hat{\mbf{i}}}^R \right \}$ and $\left \{ \mbf{I}_{\widehat{\mbf{m}}}^R, \mbf{M}_{\widehat{\mbf{m}}}^R \right \}$ for image-based and metadata-based retrievals, respectively.

\paragraph{Consistency Verifier:} The next step in the semantic integrity verification is to use these sets of evidence in combination with the query package for validation. This is performed by the consistency verifier neural network (CV). Just like MG, CV starts with first encoding the metadata, if it is structured, using the same encoder that MG uses. This allows for the encoding and semantics of metadata to be consistent across MG and CV. The CV network performs within-modality combination of query and retrieved encodings followed by cross-modality combination of information in order to assess the semantic integrity of the query package. Within-modality combination of encodings is designed as Siamese networks~\cite{koch2015siamese}, with the replicated modules termed as \textit{aggregators} (Agg) and implemented using two fully-connected neural layers. The cross-modal combination is designed as the concatenation of modality-specific information aggregates followed by a fully-connected layer. The combined information is then used to make an integrity judgement using a final fully-connected layer. This process is illustrated by Equations~\ref{eq:hii}--\ref{eq:y}.
{
\setlength{\abovedisplayskip}{10pt}
\setlength{\belowdisplayskip}{-10pt}
\setlength{\abovedisplayshortskip}{0pt}
\setlength{\belowdisplayshortskip}{0pt}
\begin{align}
    \mbf{h}^i_{\text{img}} &= \text{Agg}_{\text{img}}(\hat{\mbf{i}}, \mbf{I}_{\hat{\mbf{i}}}^R) \label{eq:hii} \\
    \mbf{h}^m_{\text{img}} &= \text{Agg}_{\text{img}}(\hat{\mbf{i}}, \mbf{I}_{\widehat{\mbf{m}}}^R) \label{eq:him} \\
    \mbf{h}^i_{\text{meta}} &= \text{Agg}_{\text{meta}}(\widehat{\mbf{m}}, \mbf{M}_{\hat{\mbf{i}}}^R) \label{eq:hmi} \\
    \mbf{h}^m_{\text{meta}} &= \text{Agg}_{\text{meta}}(\widehat{\mbf{m}}, \mbf{M}_{\widehat{\mbf{m}}}^R) \label{eq:hmm} \\
    \mbf{h}_{\text{img}} &= \text{relu} \left ( \mbf{W}_{\text{img}}^{\text{T}}[\mbf{h}^i_{\text{img}}, \mbf{h}^m_{\text{img}}] + b_{\text{img}} \right ) \label{eq:hi} \\
    \mbf{h}_{\text{meta}} &= \text{relu} \left ( \mbf{W}_{\text{meta}}^{\text{T}}[\mbf{h}^i_{\text{meta}}, \mbf{h}^m_{\text{meta}}] + b_{\text{meta}} \right ) \label{eq:hm} \\
    \mbf{h}_{\text{cross}} &= \text{relu} \left ( \mbf{W}_{\text{cross}}^{\text{T}}[\mbf{h}_{\text{img}}, \mbf{h}_{\text{meta}}] + b_{\text{cross}} \right ) \label{eq:h} \\
    y &= \sigma \left ( \mbf{W}_y^{\text{T}} \mbf{h}_{\text{cross}} + b_y  \right ) \label{eq:y}
\end{align}
}

\subsection{Training AIRD}
\label{subsec:training}

The metadata generator and the consistency verifier networks are trained adversarially using the objective described in Equation~\ref{eq:adv_obj} in a simplified notation wherein $\mbf{i}$ denotes an image and $\mbf{m}$ its real metadata.

{
\setlength{\abovedisplayskip}{-5pt}
\setlength{\belowdisplayskip}{-5pt}
\setlength{\abovedisplayshortskip}{0pt}
\setlength{\belowdisplayshortskip}{0pt}
\begin{align}
    \max_{\text{CV}} \ \min_{\text{MG}}& \ J(\text{CV}, \text{MG}) = \mathbb{E} \left [ \log \text{CV}(\mbf{i}, \mbf{m}) \right ] \nonumber \\
    &+ \mathbb{E} \left [ \log (1 - \text{CV}(\mbf{i}, \text{MG}(\mbf{i}, \mbf{m}))) \right ] \label{eq:adv_obj}
\end{align}
}

As mentioned earlier, in the case of structured metadata, the parameters of the metadata encoder are also learned jointly. However, in order to keep the training stable and the encodings consistent, the parameters of this encoder do not receive gradients directly from CV. This puts CV at a slight disadvantage by design, which encourages it to become more robust. As reflected in Equation~\ref{eq:obj}, CV is trained with real image-metadata pairs besides images with MG-generated metadata. We train the CV with two additional fake-cases, as well --- (1) $(\mbf{i}, \mbf{m}_r)$ -- image with randomly sampled fake metadata, which we call easy-negatives, and (2) $(\mbf{i}, \mbf{m}_c)$ -- image with the metadata of the most similar image-based retrieval from RD, such that $\mbf{m}_c \ne \mbf{m}$, called hard-negatives. The complete training objective is shown in Equation~\ref{eq:obj}.

{
\setlength{\abovedisplayskip}{-5pt}
\setlength{\belowdisplayskip}{-5pt}
\setlength{\abovedisplayshortskip}{0pt}
\setlength{\belowdisplayshortskip}{0pt}
\begin{flalign}
\label{eq:obj}
&\max_{\text{CV}} \ \min_{\text{MG}} \ J(\text{CV}, \text{MG})&&&& \nonumber \\
&= \mathbb{E} \left [ \log \text{CV}(\mbf{i}, \mbf{m}) \right ] + \mathbb{E} \left [ \log (1 - \text{CV}(\mbf{i}, \text{MG}(\mbf{i}, \mbf{m}))) \right ]&&&& \nonumber \\
&+ \mathbb{E} \left [ \log (1 - \text{CV}(\mbf{i}, \mbf{m}_c)) \right ] + \mathbb{E} \left [ \log (1 - \text{CV}(\mbf{i}, \mbf{m}_r)) \right ]&&&&
\raisetag{1\normalbaselineskip}
\end{flalign}
}


\section{Experimental Evaluation}
\label{sec:evaluation}

In this section, we discuss the datasets on which AIRD was evaluated, report the performance of the indexing system, provide examples of fake candidates to validate that convincing image-repurposing is indeed possible for these datasets, describe the baseline and state-of-the-art models that AIRD is compared with, and report experimental results.

\subsection{Benchmark Datasets}
\label{subsec:data}

{
\def \height {0.4\textwidth}
\begin{figure*}
\captionsetup{aboveskip=5pt,belowskip=-2pt}
\centering
\subcaptionbox{Google Landmarks\label{fig:gl_retrievals}}{%
\includegraphics[height=\height]{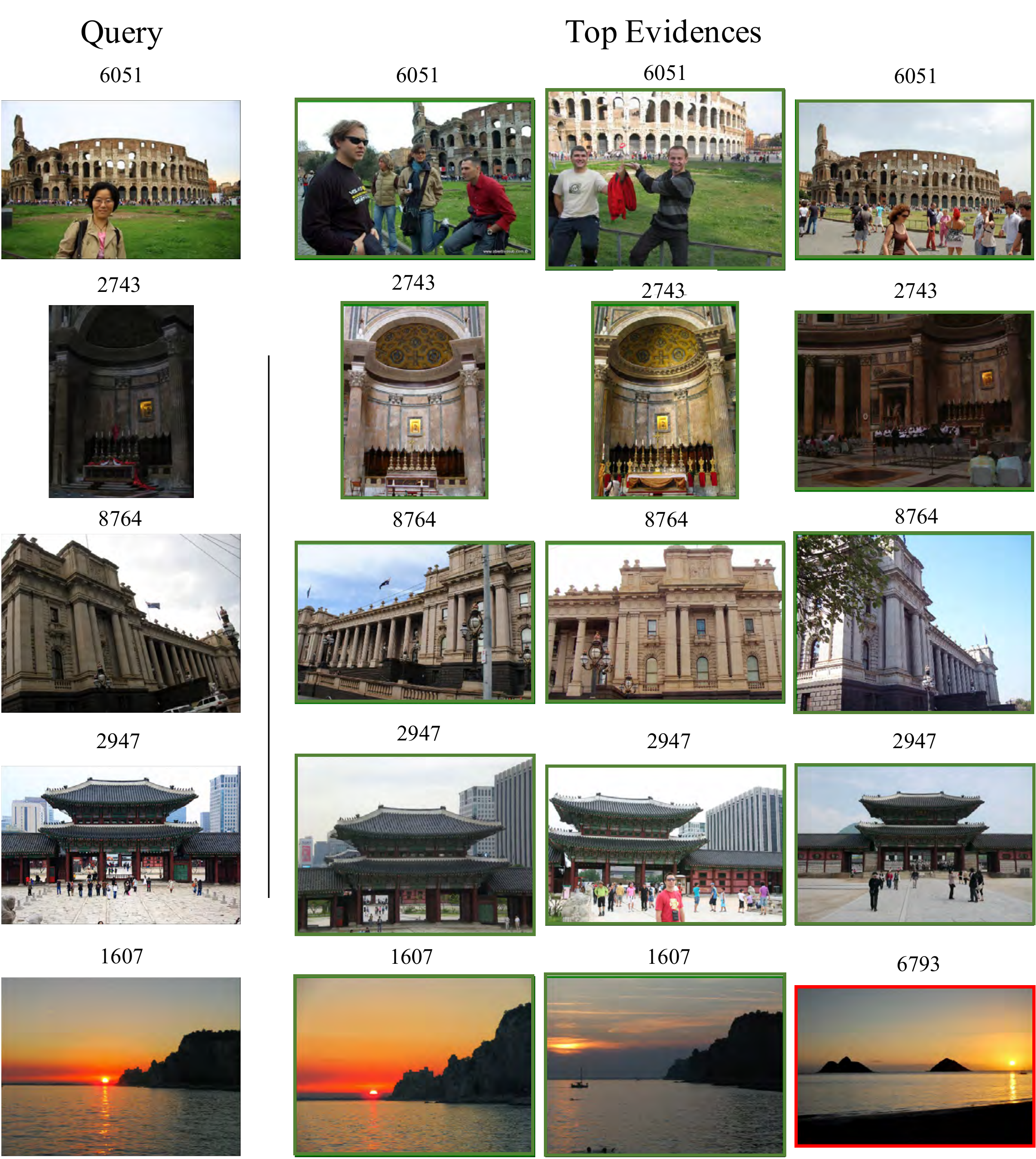}
}\hfill
\subcaptionbox{IJBC-IRD\label{fig:ijbc_retrievals}}{%
\includegraphics[height=\height]{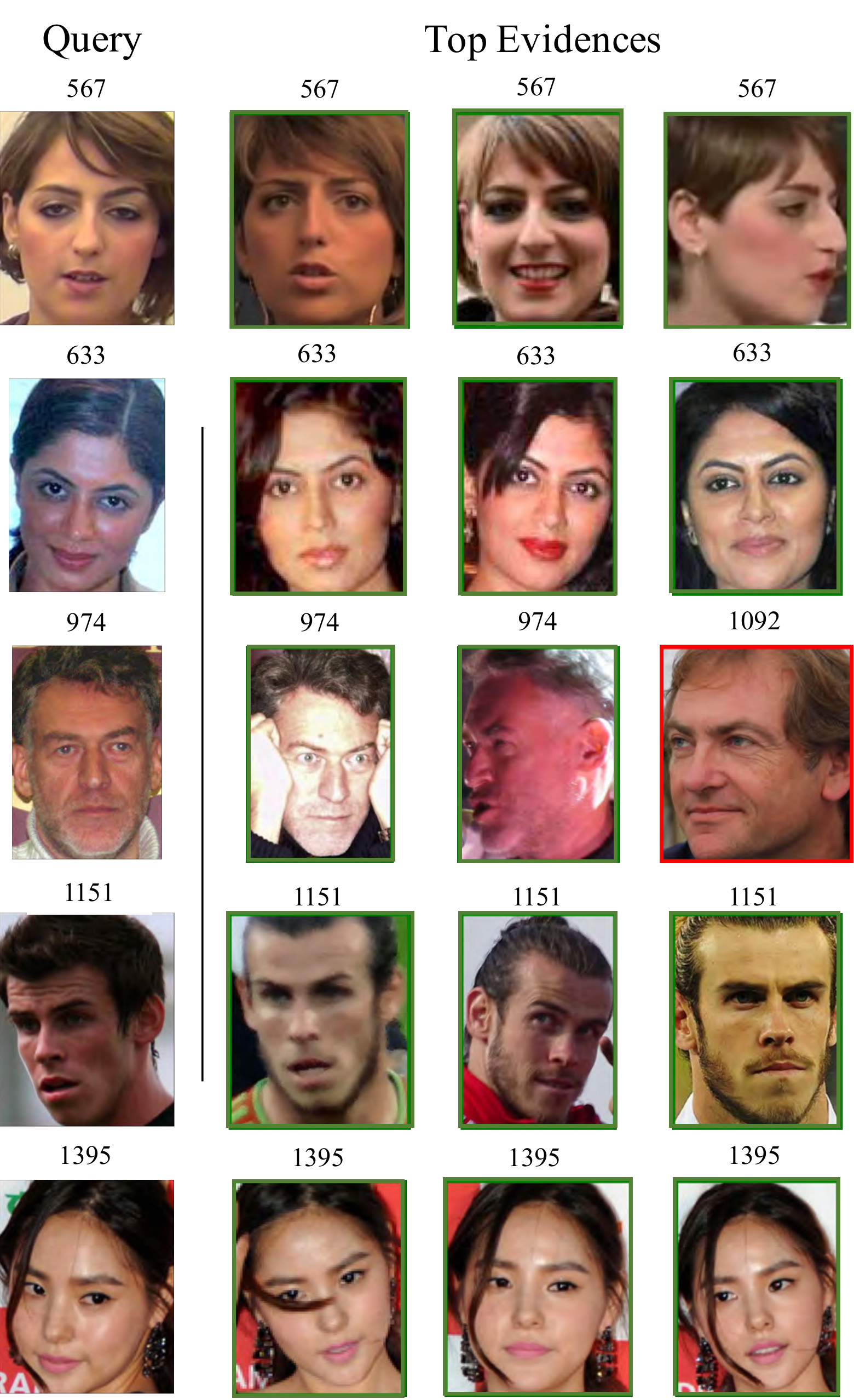}
}\hfill
\subcaptionbox{Painter by Numbers\label{fig:painters_retrievals}}{%
\includegraphics[height=\height]{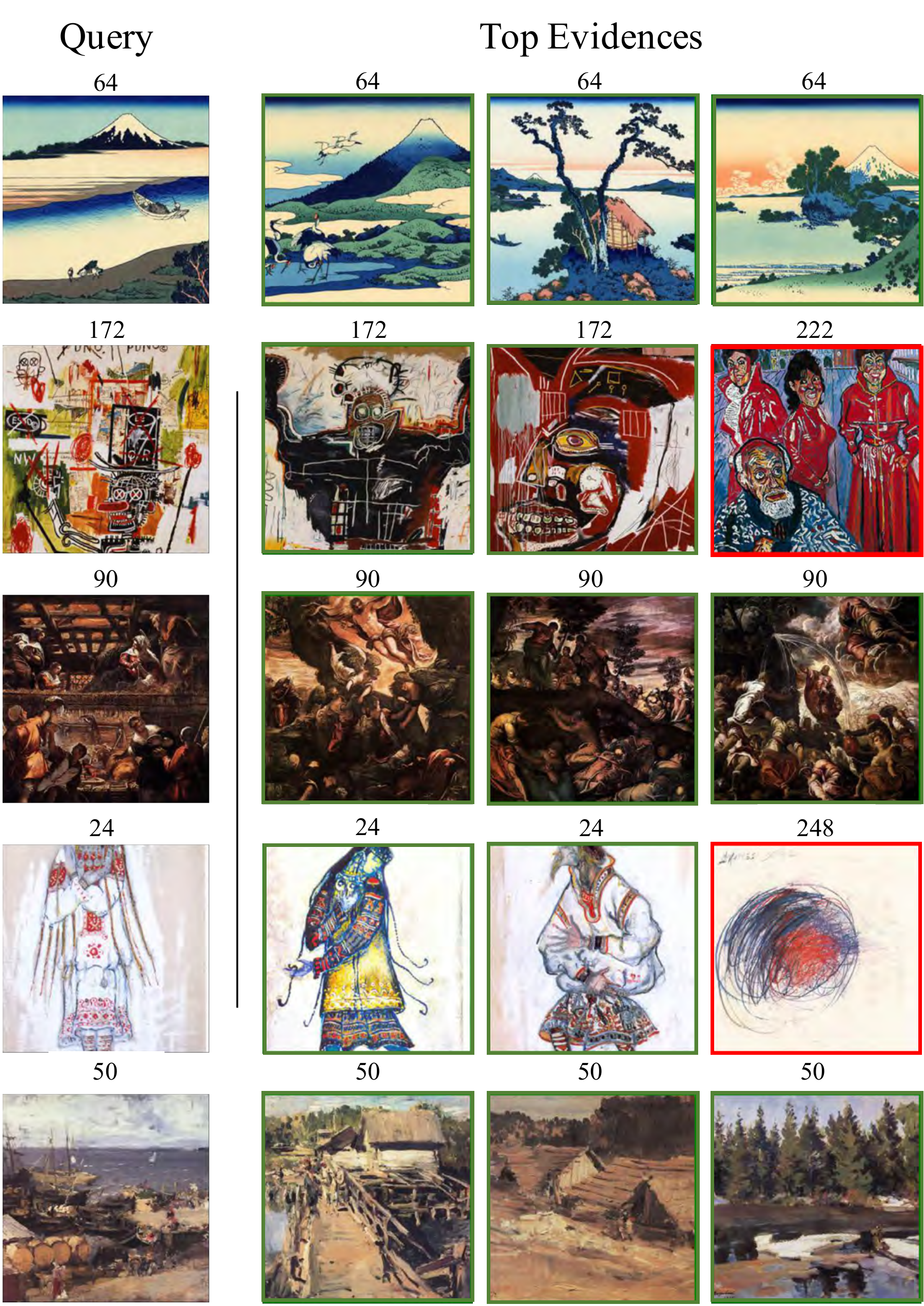}
}
\caption{Examples of image-based retrievals used by $\mathcal{D}$. The first column shows the query image and the following three are the top-3 retrievals. Each image is titled with its real metadata-identity. Correct retrievals are shown with green borders and incorrect ones with red.}
\label{fig:retrievals}
\end{figure*}
}

The proposed AIRD framework was evaluated on three diverse datasets containing specific forms of identifying metadata that are vulnerable to manipulation for image-repurposing, viz., the Google Landmarks dataset of landmark images containing location information in the form of landmark-identity, the IJBC-IRD dataset of face images containing subject-identity information, and the Painter by Numbers dataset of paintings with artist-identity metadata. Details of these datasets are provided as follows.

\paragraph{Google Landmarks:} This dataset~\cite{noh2017largescale} was released by Google for a Kaggle competition on recognizing location-landmarks from images. It is the largest worldwide dataset of images of landmarks annotated with their identity. We use this dataset for the detection of image-repurposing with location-identity manipulation, i.e., verification that the image is indeed from the claimed location. The dataset consists of 1,225,029 images spanning 14,951 different landmarks. The distribution of images across landmarks is imbalanced, with some landmarks having as low as one image while others having as high as 50,000 images. The dataset was filtered to remove landmarks with less than five images, which resulted in a total of 1,216,589 images of 13,885 landmarks. The images were encoded using a publicly available pretrained NetVLAD~\cite{bib:netvlad} model\footnote{\href{https://www.di.ens.fr/willow/research/netvlad/}{www.di.ens.fr/willow/research/netvlad/}}, which was designed and trained for place recognition, followed by dimensionality reduction with Principal Component Analysis (PCA) and $L_2$-normalization, as prescribed in~\cite{bib:netvlad}.

\paragraph{IJBC-IRD:} The IARPA Janus Benchmark C (IJB-C)~\cite{maze2018iarpa}\footnote{\href{https://www.nist.gov/programs-projects/face-challenges}{www.nist.gov/programs-projects/face-challenges}} dataset is a novel face recognition benchmark. It has tough variability with face imagery presenting a wide range of poses, harsh illuminations, occlusions, aging and other challenging conditions. For all these reasons, the series IJB--\{A,B,C\}~\cite{Klare_2015_CVPR,Whitelam_2017_CVPR_Workshops,maze2018iarpa} quickly became the \emph{de facto} standard for face recognition in the wild.

With these motivations, seeking realistic scenarios for face repurposing detection, we selected a subset of IJB-C to create a new benchmark, dubbed ``IJB-C Image Repurposing Detection'' (IJBC-IRD). IJBC-IRD shares the same media as IJB-C but focuses on the subjects  that are more likely to be used for face identity repurposing. To do so, we favored subjects with ample intra-class variations (picking individuals with at least five media from the IJB-C metadata) and only considering \emph{still images}, thus discarding all the motion frames. We motivate the use of still images, since we argue that clean, good quality images serve better for the face repurposing task -- frames from videos usually contain motion blur and lack of discriminative facial features, making the impersonification of a subject less believable.

The IJBC-IRD dataset contains 16,377 images spanning 1,649 subjects. We employ a state-of-the-art face recognition system to encode face images by following the procedure of~\cite{masi2017rapid,masi2016we}. We chose this system for its performance and for its pose-invariance capability~\cite{Masi:18:learning}. The face encoder is a single convolutional neural network based on a deep residual architecture, following the same training procedure described in~\cite{chang17fpn}.
Faces are encoded using the activations of the penultimate layer and the descriptors are decorrelated via PCA and signed square rooting. The final encoding for an image is the result of average-pooling estimated views rendered at different angles with the original 2D aligned image. In general, we use the same recognition pipeline from~\cite{chang17fpn} and we refer to that for more details.

\paragraph{Painter by Numbers:} This dataset~\cite{painter_by_numbers} was created for a Kaggle competition to determine whether pairs of paintings belonged to the same artist, in order to develop technologies for detecting art forgeries. We use this dataset to evaluate the detection of image repurposing where a painting's artist-ownership has been manipulated, i.e., detecting whether the painting was indeed painted by the claimed artist. The dataset contains 103,250 images from 2,319 different artists. Just like previous cases, this dataset is imbalanced with frequency ranging from one painting to 500 paintings per artist. The dataset was filtered to pick images of paintings of the top 1,000 most frequent artists in the dataset resulting in 72,863 paintings. The images were encoded using the model\footnote{\href{https://github.com/inejc/painters}{www.github.com/inejc/painters}} that won the competition, followed by $L_2$-normalization.

\paragraph{}All datasets were split into training and testing sets containing $80\%$ and $20\%$ images, respectively, using stratified sampling. The training splits of the datasets were also additionally used as reference datasets in all the experiments. All possible retrievals in the form of fake candidates for the counterfeiter and evidence for the detector were precomputed in order to speed up the training process.

{
\def \height {0.4\textwidth}
\begin{figure*}
\captionsetup{aboveskip=5pt,belowskip=-2pt}
\centering
\subcaptionbox{Google Landmarks\label{fig:gl_fake}}{%
\includegraphics[height=\height]{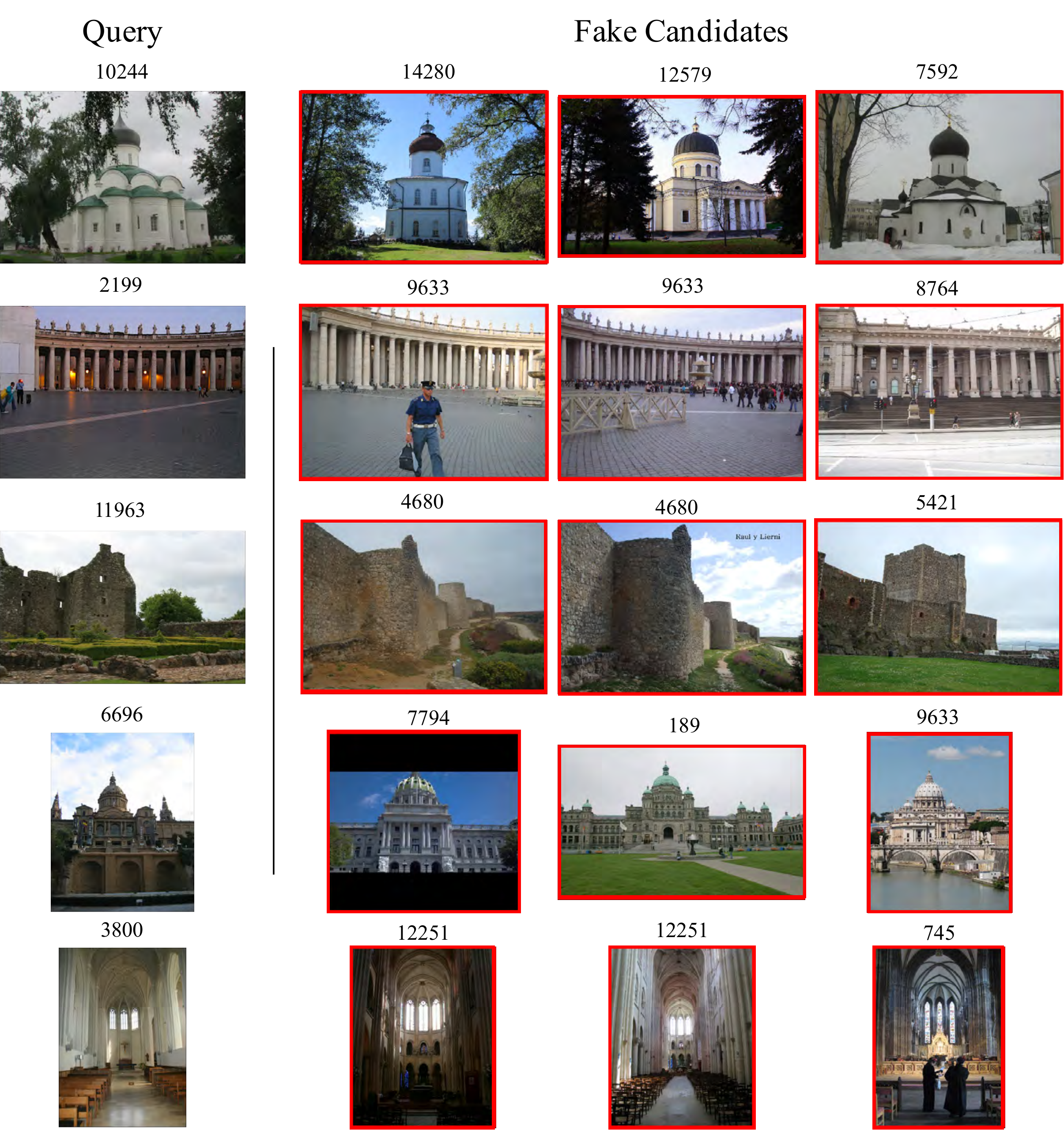}
}\hfill
\subcaptionbox{IJBC-IRD\label{fig:ijbc_fake}}{%
\includegraphics[height=\height]{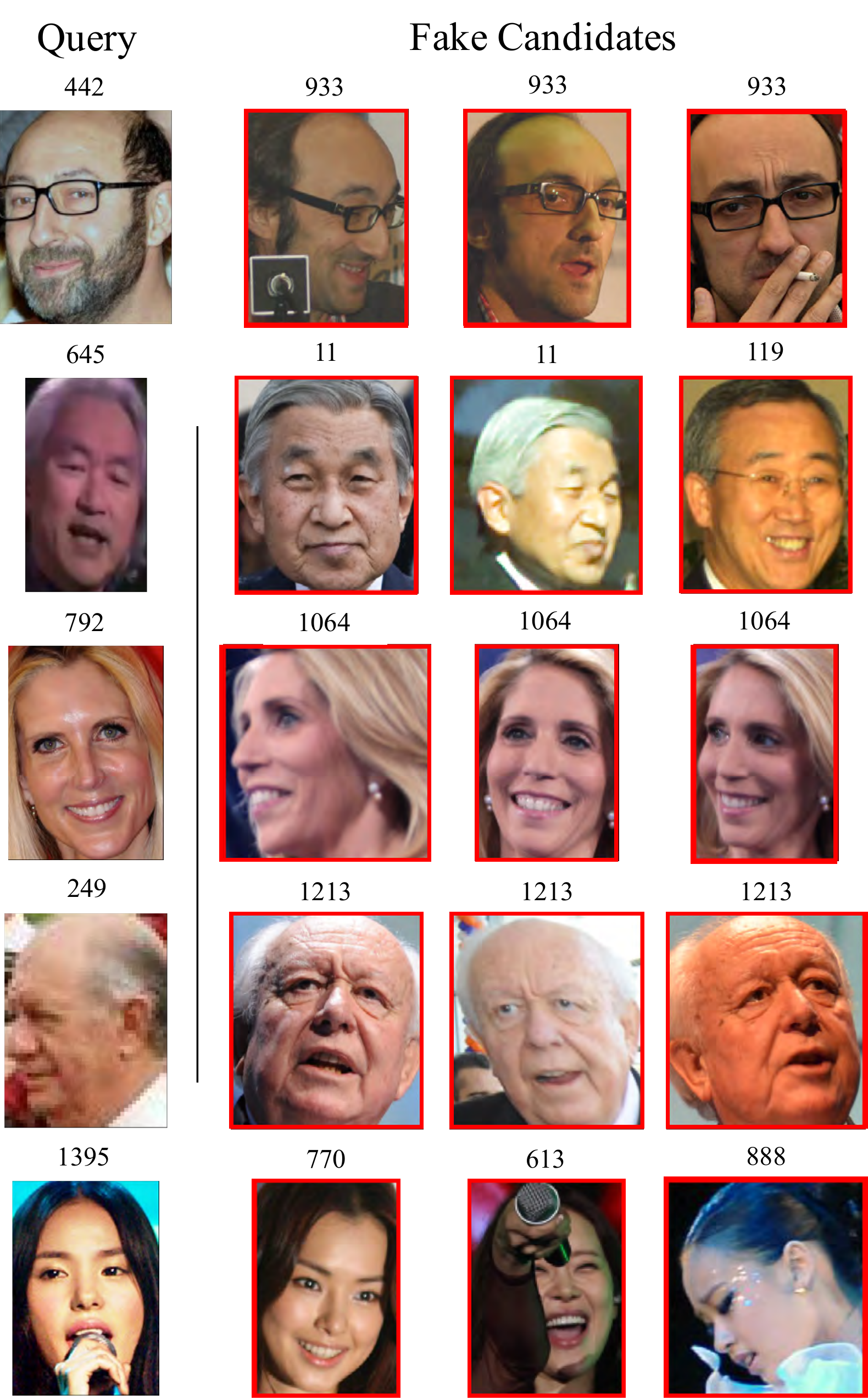}
}\hfill
\subcaptionbox{Painter by Numbers\label{fig:painters_fake}}{%
\includegraphics[height=\height]{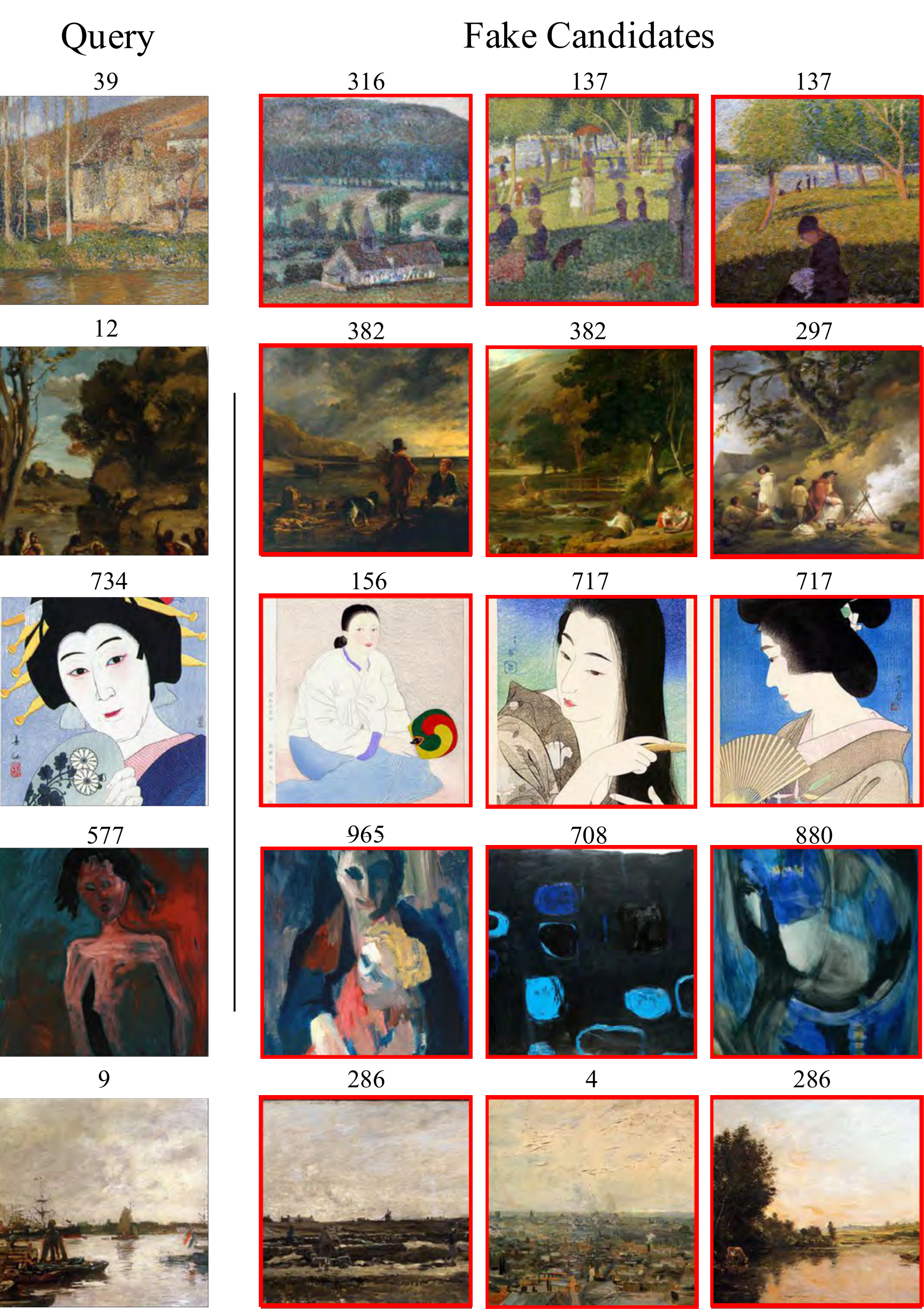}
}
\caption{Examples of fake-candidates used by $\mathcal{C}$. The first column shows the query image and the following three are fake-candidates.}
\label{fig:fake}
\end{figure*}
}

\subsection{Indexing and Retrieval Performance}

Figures~\ref{fig:gl_retrievals}, \ref{fig:ijbc_retrievals} and \ref{fig:painters_retrievals} show qualitative results of image-based retrievals for the Google Landmarks, IJBC-IRD and Painter by Numbers datasets, respectively. In each of these figures, the first column shows query images, followed by the top three image-based retrievals from the reference dataset. In the figures, we use green borders to show cases where the metadata of the retrieved image matches that of the query and red ones to show cases where it does not. The results show that the retrieval system returns images that are very similar to the query image, with usually the same metadata. However, it also makes mistakes sometimes when there are very similar images in the reference dataset that have dissimilar metadata. This can be attributed to the cascading effect of errors in the semantic encoding of images followed by approximation errors in the nearest neighbor search.

The Mean Average Precision at $K$ (MAP@K) and Precision at $K$ (Precision@K) with respect to accurate matching of metadata between query and retrieved images are reported in Table~\ref{tab:retrievals} to quantify the performance of the retrieval system on the aforementioned datasets. The results show that the retrieval performance is the best on IJBC-IRD, followed closely by Google Landmarks and relatively lower on Painter by Numbers, which is an especially challenging dataset because its metadata (the painting's artist) is more subtle than that of the other datasets (subject or location identity).

{
\setlength{\abovedisplayskip}{-5pt}
\setlength{\belowdisplayskip}{-10pt}
\setlength{\abovedisplayshortskip}{0pt}
\setlength{\belowdisplayshortskip}{0pt}
\begin{table}[t]
\centering
\caption{\label{tab:retrievals}Performance of image-based retrieval in similar search of images with the same metadata value. Metrics are reported only for $K$=3, which is the number of packages retrieved in our experiments. GL stands for Google Landmarks and PbN for Painter by Numbers.}
\small
\begin{tabular}{l ccc}
\toprule
\textbf{Metric} & \textbf{GL}  & \textbf{IJBC-IRD} & \textbf{PbN} \\ 
\cmidrule(r){1-1}  \cmidrule(l){2-4}
MAP@3 & 0.8127 & 0.8396 & 0.6147 \\
Precision@3 & 0.8404 & 0.8537 & 0.6326 \\
\bottomrule
\end{tabular}
\end{table}
}

\subsection{Fake Candidates for Counterfeiter}

The success of the proposed AIRD framework and the adversarial training, more specifically, relies on the ability of the counterfeiter to find convincing fake candidates. We present samples of fake candidates used by $\mathcal{C}$ for the Google Landmarks, IJBC-IRD, and Painter by Numbers datasets in Figures~\ref{fig:gl_fake}, \ref{fig:ijbc_fake} and \ref{fig:painters_fake}, respectively. The first column of each of these figures shows the image selected by the counterfeiter to repurpose. The following three columns show fake candidates. The results show that these datasets contain very convincing similar images with dissimilar metadata such that one could be confused for the other. \emph{It is also important to note that in real cases of image repurposing, the audience does not see the query image (the first column) as a part of the information package (e.g., a fake news article) but only one of the fake candidates.} This makes it easy to fool people.

\subsection{Baseline and State-of-the-Art Models}

The proposed AIRD framework is designed to make additional information available to the detector in the form of retrieval from the reference dataset. However, this setup also allows for the development of several non-learning methods involving direct comparison of query and retrieved packages using similarity metrics to make integrity assessments. We discuss these models below and use them as baselines for comparing the proposed AIRD framework with:
\begin{itemize}
    \item $B_1$ -- similarity between query metadata and that of query-image-based retrievals from the reference dataset
    \item $B_2$ -- similarity between query image and images retrieved from the reference dataset using query metadata
    \item $B_3$ -- similarity between images retrieved using query image and those retrieved using query metadata
    \item $B_4$ -- similarity between metadata of query-image-based retrievals and those retrieved using metadata
\end{itemize}

Another method of metadata validation is the use of a metadata-predictor (MP) model. Given a query image and its metadata, MP first predicts the metadata for the image and then matches it with the claimed metadata. If the two match, MP tags the query package as valid.

Previous works in this domain~\cite{bib:jaiswal2017, bib:sabir2018} focused on the detection of metadata modalities that are continuous in nature, such as captions and GPS coordinates in the form of latent encodings. The method of~\cite{bib:jaiswal2017} is not suitable for structured metadata because it relies on learning a joint representation of images and captions. The publicly available deep multitask model (DMM) of~\cite{bib:sabir2018}\footnote{\href{https://www.github.com/Ekraam/MEIR}{www.github.com/Ekraam/MEIR}} was evaluated and we report the scores of this model on the IJBC-IRD and Painter by Numbers datasets. The package-similarity based retrieval used in their framework was not feasible on the Google Landmarks datasets, which has 1.2 million images. Hence, DMM could not be evaluated on this dataset.

In addition, a non-adversarial version of AIRD is evaluated as an ablation study. We call this model the non-adversarial detector (NAD). NAD is trained with real image-metadata pairs along with $(\mbf{i}, \mbf{m}_r)$ and $(\mbf{i}, \mbf{m}_c)$ for easy and hard negatives, respectively, as described in Section~\ref{subsec:training}.

\subsection{Results}

In order to evaluate the proposed framework and the aforementioned models, we use $K=3$, i.e., both the counterfeiter and the detector retrieve three packages from the RD as fake candidates and evidence, respectively. The similarity threshold for making decisions with the non-learning baseline models $B_1$, $B_2$ and $B_3$ were tuned on the training dataset. The datasets used in the experimental evaluation contain structured metadata. Hence, $B_4$ was not evaluated because it reduces to $B_1$. MP was implemented as a three-layer fully connected neural network trained with the same encodings that AIRD was trained on, which were generated using dedicated deep neural networks as described in Section~\ref{subsec:data}.

Following the approach of previous works~\cite{bib:jaiswal2017,bib:sabir2018}, $F_1$-tampered ($F_1$-tamp; calculated by treating $y=\text{fake}$ as the positive class), $F_1$-clean (calculated with $y=\text{real}$ as the positive class), and Area Under Receiver Operating Curve (AUC) were used to quantify the performance of the models. We also report the accuracy scores (ACC) as an additional metric of model performance. All the models were tested on real image-metadata pairs as well as $(\mbf{i}, \mbf{m}_r)$ pairs of images with randomly sampled fake metadata, following the evaluation methodology of previous works~\cite{bib:jaiswal2017,bib:sabir2018}. The models were additionally evaluated on hard-negatives $(\mbf{i}, \mbf{m}_c)$.

Tables~\ref{tab:gl}, \ref{tab:ijbc} and \ref{tab:painters} present results of the experiments. The results show that the proposed AIRD framework outperforms all other models on all metrics. While the non-adversarial detector (NAD) performs better than other baseline models, its performance is inferior to the complete AIRD. This additional boost in performance is, therefore, credited to the adversarial training of the detector with the counterfeiter. The proposed framework outperforms the prior state-of-the-art DMM model by a large margin, showing that DMM is not suitable for the case of image repurposing where the metadata comprises structured identity information.

{
\setlength{\belowdisplayskip}{0pt}
\setlength{\belowdisplayshortskip}{0pt}
\addtolength{\tabcolsep}{-0.25pt}
\begin{table}[t]
\captionsetup{skip=2pt}
\centering
\caption{\label{tab:gl}Evaluation Results on Google Landmarks Dataset.}
\small
\begin{tabular}{l cccccc}
\toprule
\textbf{Metric} & $\boldsymbol{B_1 (B_4)}$ & $\boldsymbol{B_2}$ & $\boldsymbol{B_3}$  & \textbf{MP} & \textbf{NAD} & \textbf{AIRD} \\
\cmidrule(r){1-1}  \cmidrule(l){2-7}
 $F_1$-tamp & 0.91 & 0.81 & 0.81  & 0.88 & 0.91 & \textbf{0.95} \\
$F_1$-clean & 0.81 & 0.37 & 0.39  & 0.87 & 0.90 & \textbf{0.91} \\
$ACC$ & 0.86 & 0.72 & 0.71  & 0.88 & 0.90 & \textbf{0.94} \\
$AUC$ & 0.88 & 0.79 & 0.76  & 0.94 & 0.95 & \textbf{0.98} \\
\bottomrule
\end{tabular}
\end{table}
\addtolength{\tabcolsep}{0.25pt}
}

\addtolength{\tabcolsep}{-2.5pt}
\begin{table}[t]
\captionsetup{skip=2pt}
\centering
\caption{\label{tab:ijbc}Evaluation Results on IJBC-IRD Dataset.}
\small
\begin{tabular}{l ccccccc}
\toprule
 \textbf{Metric} & $\boldsymbol{B_1 (B_4)}$ & $\boldsymbol{B_2}$ & $\boldsymbol{B_3}$  & \textbf{MP} & \textbf{DMM} & \textbf{NAD} & \textbf{AIRD} \\
\cmidrule(r){1-1}  \cmidrule(l){2-8}
  $F_1$-tamp & 0.91 & 0.90 & 0.90  & 0.91 & 0.50 & 0.93 & \textbf{0.95} \\
  $F_1$-clean & 0.83 & 0.75 & 0.77  & 0.84 & 0.72 & 0.86 & \textbf{0.89} \\
  $ACC$ & 0.89 & 0.86 & 0.87  & 0.89 & 0.65 & 0.90 & \textbf{0.93} \\
  $AUC$ & 0.90 & 0.93 & 0.92  & 0.94 & 0.76 & 0.95 & \textbf{0.97} \\
\bottomrule
\end{tabular}
\end{table}
\addtolength{\tabcolsep}{2.5pt}

\addtolength{\tabcolsep}{-2.5pt}
\begin{table}[t]
\captionsetup{skip=2pt}
\centering
\caption{\label{tab:painters}Evaluation Results on Painter by Numbers Dataset.}
\small
\begin{tabular}{l ccccccc}
\toprule
 \textbf{Metric} & $\boldsymbol{B_1 (B_4)}$ & $\boldsymbol{B_2}$ & $\boldsymbol{B_3}$  & \textbf{MP} & \textbf{DMM} & \textbf{NAD} & \textbf{AIRD} \\
\cmidrule(r){1-1}  \cmidrule(l){2-8}
  $F_1$-tamp & 0.81 & 0.80 & 0.80  & 0.76 & 0.22 & 0.82 & \textbf{0.83} \\
  $F_1$-clean & 0.46 & 0.16 & 0.18  & 0.58 & 0.64 & 0.63 & \textbf{0.68} \\
  $ACC$ & 0.72 & 0.68 & 0.69  & 0.69 & 0.51 & 0.76 & \textbf{0.77} \\
  $AUC$ & 0.61 & 0.77 & 0.71  & 0.79 & 0.53 & 0.80 & \textbf{0.84} \\
\bottomrule
\end{tabular}
\end{table}
\addtolength{\tabcolsep}{2.5pt}


\section{Conclusion}
\label{sec:conclusion}

We presented a novel framework for image repurposing detection that is modeled after the real-world adversarial interplay between nefarious actors who spread misinformation and watchdogs who verify information. The proposed framework is composed of a counterfeiter and a detector, which are trained adversarially. Like real-world, both the models have access to world knowledge through retrieval of information from a reference dataset, which they use to their advantage. We described the model components along with the training strategy. The framework was evaluated on the Google Landmarks dataset with location-identiy, IJBC-IRD with subject-identity and Painter by Numbers dataset with painting-artist as metadata. Results show that the proposed framework outperforms all baseline models and prior state-of-the-art on all metrics on a diverse collection of datasets.


\section*{Acknowledgements}
This work is based on research sponsored by the Defense Advanced Research Projects Agency under agreement number FA8750-16-2-0204. The U.S. Government is authorized to reproduce and distribute reprints for governmental purposes  notwithstanding any copyright notation thereon. The views and conclusions contained herein are those of the authors and should not be interpreted as necessarily representing the official policies or endorsements, either expressed or implied, of the Defense Advanced Research Projects Agency or the U.S. Government.

{\small
\bibliographystyle{ieee_fullname}
\bibliography{main}
}

\end{document}